\documentclass[12pt,a4paper]{article}
\usepackage[english]{babel}
\usepackage[utf8]{inputenc}
\usepackage{amsmath}
\usepackage{amsfonts}
\usepackage{amssymb}
\usepackage{biblatex}[=3.2]
\usepackage[left=2.5cm,right=2.5cm,top=2.5cm,bottom=2.5cm]{geometry}
\usepackage{graphicx}
\usepackage{float} 
\usepackage{array}
\usepackage{caption}
\usepackage{listings}
\usepackage{xcolor}
\usepackage{hyperref}
\usepackage{csquotes}

\begin{document}

\title{Impact of Sampling Techniques and Data Leakage on XGBoost Performance in Credit Card Fraud Detection}

\author{Siyaxolisa Kabane\\
Department of Computer Science\\
University of Fort Hare\\
\href{mailto:202378066@ufh.ac.za}{202378066@ufh.ac.za}}
\date{}

\maketitle

\begin{abstract}
Credit card fraud detection remains a critical challenge in financial security, with machine learning models like XGBoost(eXtreme gradient boosting) emerging as powerful tools for identifying fraudulent transactions. However, the inherent class imbalance in credit card transaction datasets poses significant challenges for model performance. Although sampling techniques are commonly used to address this imbalance, their implementation sometimes precedes the train-test split, potentially introducing data leakage.

This study presents a comparative analysis of XGBoost's performance in credit card fraud detection under three scenarios: Firstly without any imbalance handling techniques, secondly with sampling techniques applied only to the training set after the train-test split, and third with sampling techniques applied before the train-test split. We utilized a dataset from Kaggle of 284,807 credit card transactions, containing 0.172\% fraudulent cases, to evaluate these approaches.

Our findings show that although sampling strategies enhance model performance, the reliability of results is greatly impacted by when they are applied. Due to a data leakage issue that frequently occurs in machine learning models during the sampling phase, XGBoost models trained on data where sampling was applied prior to the train-test split may have displayed artificially inflated performance metrics. Surprisingly, models trained with sampling techniques applied solely to the training set demonstrated significantly lower results than those with pre-split sampling, all the while preserving the integrity of the evaluation process.

This study raises concerns about the necessity of pre-split sampling, which risks data leakage, and emphasizes the vital significance of using appropriate sampling techniques in credit card fraud detection models. According to our findings, sampling approaches can improve model performance sufficiently without sacrificing the validity of outputs when properly applied just to the training data. These observations aid in the development of improved procedures for managing unbalanced datasets for fraud detection and related classification issues in applications using financial machine learning.
\end{abstract}

\textbf{Keywords:} Machine Learning, Credit Card Fraud Detection, XGBoost, Class Imbalance, Data Leakage 

\section{Introduction}

In this digital age, e-Commerce has emerged as a vital channel for global business transactions, leading to an increase in credit card usage for online purchases.This rise in credit card transactions has undoubtedly made fraud a lucrative endeavor for cybercriminals. Credit card fraud in countries like South Africa is on the rise due to technological advances and identity theft, resulting in significant financial losses of about billion Rands for both card companies and consumers\cite{Budhram2016Lost}. Fraudsters are employing increasingly advanced methods, posing significant challenges to financial institutions and credit card issuers. Fraud encompasses any deliberate act of deception, misrepresentation, or concealment of material facts undertaken with the intention of securing unauthorized financial gain, tangible assets, or personal advantages at the expense of others. This behavior may come through various schemes, ranging from simple misrepresentations to elaborate criminal enterprises, all sharing the common element of intentional deceit for unlawful profit. The unlawful use of credit card information for digital or physical purchases is called credit card fraud. Since cardholders often submit their card number, expiration date, and card verification number over the phone or online, fraud can occur during digital transactions. With relatively little experience in fraud detection, most auditors find it challenging to effectively identify fraudulent activities \cite{JOHNSON1993467}, leading us to machine learning approaches for more reliable results\cite{9999220}.

The detection of credit card fraud has traditionally relied on rule-based systems, where predetermined rules are used to flag potentially fraudulent transactions, however these systems struggle with high false positive rates and cannot easily adapt to new types of fraud\cite{ABDALLAH201690}. Consequently, machine learning and data mining techniques have gained traction as alternatives or complements to rule-based methods, allowing for more adaptive and accurate detection of fraudulent transactions \cite{bhattacharyya2011data, niu2019comparisonstudycreditcard}. Among the machine learning models, ensemble methods such as Random Forest and gradient boosting models like XGBoost have demonstrated significant promise due to their ability to capture complex patterns in data and improve detection accuracy \cite{breiman2001random, chen2016xgboost,article}.

One of the major challenges in the detection of credit card fraud lies in the inherent class imbalance in transaction datasets, where fraudulent transactions constitute a very small proportion of total transactions. This imbalance can bias models towards the majority (non-fraudulent) class, reducing their effectiveness in identifying fraud. Various techniques, such as Synthetic Minority Over-sampling Technique (SMOTE),random over-sampling,under-sampling \cite{yang2024impact} and synthetic data generation via generative adversarial networks \cite{lee2021gan}, are commonly applied to mitigate this issue by either generating synthetic fraudulent samples or reducing the number of non-fraudulent transactions \cite{chawla2002smote}. These methods are not without pitfalls though; if sampling is applied before the data split into training and test sets, data leakage may occur, compromising the evaluation of the model’s performance \cite{rosenblatt2024data_leakage}.Data leakage occurs when information from the test set inadvertently enters the training process, often due to improperly applied sampling techniques or pre-processing steps performed before the train-test split. This issue leads to artificially inflated accuracy and precision metrics, as the model has access to data it would not encounter in a real-world scenario, rendering the evaluation results unreliable\cite{samala2020hazards_data_leakage}.

XGBoost, a gradient boosting model designed for speed and performance, is particularly suited for high-dimensional and structured data, making it a popular choice in credit card fraud detection applications \cite{chen2016xgboost}. Despite its effectiveness, XGBoost’s performance can be substantially affected by the class imbalance and potential data leakage, necessitating careful handling of data preprocessing and sampling techniques.

The present study will attempt to analyze the effect of different sampling strategies on XGBoost performance in fraud detection of credit cards and find out consequences of data leakage when sampling is done before and after the train-test split. By comparing model performances under these different scenarios, this work hopes to derive best practices that preserve model evaluation validity. It also contributes to the growing literature on credit card fraud detection methodologies by underlining the need for rigorous and leak-proof assessment methods in order to ensure practical financial applications of the machine learning models.

\section{Related Works}
Credit card fraud detection is considered an heavily studied field, in the realm of machine learning because of the complexity involved in spotting instances of fraudulent transactions among a large volume of legitimate ones. The imbalance between transactions and fraudulent ones presents an obstacle, for traditional machine learning methods. Researchers have come up with solutions to tackle these challenges. From using sampling methods to balance class distributions to creating innovative generative models that mimic authentic fraud patterns and developing robust anomaly detection techniques to detect subtle deviations, in regular transaction behaviors effectively. These strategies have evolved over time to combat fraud tactics while keeping false positives at bay and ensuring smooth processing of legitimate customer transactions.

Alamri and Ykhlef~\cite{alamri2022survey} provide an extensive survey on credit card fraud detection, emphasizing sampling techniques. Their analysis reveals that methods like oversampling, undersampling, and hybrid approaches effectively manage class imbalance but require careful implementation to avoid pitfalls like data leakage. Sampling techniques are vital in ensuring a balanced dataset, but as highlighted by Rosenblatt et al.~\cite{rosenblatt2024data_leakage} and Samala et al.~\cite{samala2020hazards_data_leakage}, improper application can lead to data leakage.The significance of strict processes to preserve assessment integrity is shown by these research, which show that data leakage can inflate performance measurements by enabling test data to affect model training.

Generative Adversarial Networks (GANs) and other generative models have emerged as powerful tools for addressing data imbalance challenges. Lee and Park~\cite{lee2021gan} pioneered a GAN-based framework for intrusion detection that strategically generates synthetic samples of minority classes, leading to significant improvements in detection accuracy. Building on this concept, Choi et al.~\cite{choi2017generating} demonstrated GANs' remarkable capability to generate highly realistic, multi-label healthcare records while preserving complex relationships between features. Their success in healthcare data generation offers compelling evidence for GANs' potential in financial fraud detection, where similar requirements for maintaining realistic feature correlations and data patterns are crucial. This as an area we will also explore in our techniques. 

Semi-supervised anomaly detection methods have also been explored. Wulsin et al.~\cite{wulsin2010semi} presented a semi-supervised anomaly detection technique using deep belief networks to detect anomalies in EEG data.This method demonstrates how semi-supervised models may spot minute patterns in unbalanced data, providing information about how to use them to fraud detection.

In banking, traditional machine learning techniques remain relevant. Hashemi et al.~\cite{hashemi2023fraud} analyzed various models for fraud detection, noting the effectiveness of ensemble methods such as XGBoost and Random Forests. Their study stresses the need for rigorous model evaluation, as improper techniques can lead to inflated results due to data leakage.A paper focusing on the XGBoost model and SOMTE was written by Qasim et al.~\cite{9719580}. 

Taken together, these studies outline the critical role that sampling strategies, generative models, and rigorous evaluation protocols play in improving credit card fraud detection, along with data leakage mitigation to protect the integrity and generalizability of model performance.

\section{XGBoost Model Overview}

The well-known ensemble machine learning algorithm XGBoost (eXtreme Gradient Boosting)\cite{chen2016xgboost} was created for applications involving regression and classification. In order to generate a powerful predictive model, it integrates several weak learners, usually decision trees. Efficiency, scalability, and the capacity to handle high-dimensional datasets are XGBoost's main advantages; these are especially helpful in domains such as credit card fraud detection.

\subsection{Gradient Boosting in XGBoost}

XGBoost uses a specialized form of gradient boosting to optimize model performance. At each iteration, the algorithm calculates the gradient and second-order gradient (Hessian) of the loss function for each data point. This process helps to direct the model's updates toward minimizing residual errors, resulting in more accurate predictions. Additionally, XGBoost adjusts the weights of each leaf node in the decision trees, balancing prediction accuracy with regularization to prevent overfitting and enhance generalization.

Several features set XGBoost’s gradient boosting apart:
\begin{itemize}
    \item \textbf{Second-order Gradient Optimization}: By leveraging both first and second-order derivatives, XGBoost can make more precise adjustments, improving model accuracy and convergence speed.
    \item \textbf{Regularization}: XGBoost applies L1 (Lasso) and L2 (Ridge) regularization to penalize overly complex models, thus helping to avoid overfitting and ensure that the model generalizes well.
    \item \textbf{Handling Missing Values}: XGBoost has a built-in capability to handle missing data by learning optimal directions for missing values during tree construction.
    \item \textbf{Parallel and Distributed Computation}: XGBoost supports parallel tree construction, allowing it to process large datasets efficiently across multiple cores or machines.
\end{itemize}

\subsection{Handling Imbalanced Data with XGBoost}

In applications like fraud detection, where there is a class imbalance, XGBoost can be tuned to focus more on minority classes. This is achieved by adjusting parameters(e.g scale pos weight) to give more importance to the positive (fraudulent) class. Additionally, oversampling and undersampling techniques, like SMOTE and random undersampling, are often applied to balance the dataset before training the model, thereby enhancing its sensitivity to fraudulent cases.

\subsection{Preventing Data Leakage in XGBoost}

Data leakage is a significant concern when handling imbalanced datasets in machine learning. Leakage can occur if sampling techniques are applied before the data split into training and test sets, allowing the model to gain information from the test set inadvertently. In this study, sampling techniques were applied exclusively to the training set after the split to ensure a fair evaluation and avoid inflated performance metrics.

\section{Sampling Techniques}
In order to deal with the high class imbalance found in credit card fraud detection datasets, this research and other studies we will take a look at have employed a number of sampling methodologies that include Random Over-Sampling, Random Under-Sampling, the Synthetic Minority Over-sampling Technique (SMOTE), and Conditional Generative Adversarial Networks (CGAN) for synthetic data generation. Application of such methodologies leads to sensitivity rise of the model toward the minority (fraudulent) class, raising the general accuracy of detection and reducing the possibility of overfitting toward the majority class.

\subsection{Random Over-Sampling and Under-Sampling}
Random Over-Sampling (ROS) involves duplicating samples from the minority class to balance the dataset\cite{10180891}, while Random Under-Sampling (RUS) reduces the majority class by randomly removing samples. These techniques help mitigate class imbalance by adjusting the dataset distribution, thereby making the model more sensitive to fraudulent cases. However, excessive oversampling or undersampling may lead to overfitting or loss of valuable information from the majority class, respectively\cite{yang2024impact}.

\subsection{Synthetic Minority Over-sampling Technique (SMOTE)}
SMOTE \cite{Chawla2002} is an advanced over-sampling technique that generates synthetic samples for the minority class by interpolating between existing minority class instances. By creating plausible new data points rather than duplicating existing ones, SMOTE mitigates overfitting and allows the model to better generalize to unseen data.

\subsection{Conditional Generative Adversarial Networks (CGAN)}
To further enhance the dataset, CGANs were utilized to generate realistic synthetic data for the minority class. Unlike traditional over-sampling methods, CGANs use deep learning techniques to capture the complex distributions of minority samples, producing highly realistic synthetic data points that maintain feature correlations. CGANs have been shown to effectively reduce class imbalance while preserving data characteristics essential for accurate fraud detection \cite{lee2021gan}. This technique provides a robust alternative to traditional sampling, at times improving model performance on imbalanced datasets \cite{DOUZAS2018464}.

Applying these sampling techniques only to the training set mitigated the risk of data leakage and ensured that model evaluation remained fair and unbiased, as detailed in related works \cite{rosenblatt2024data_leakage}.

\section{Tools Used}

\begin{itemize}
    \item \textbf{Google Colab}:The cloud-based platform Google Colab, which offers a free environment with GPU access, was used for all calculations for our experiments. For the vast data processing and model training requirements of this study, Colab was perfect since it made it possible to handle resource-intensive machine learning operations efficiently.
    
    \item \textbf{Python and Jupyter Notebooks}:Python was the main programming language utilized, and it was implemented in Google Colab's Jupyter Notebooks. A systematic workflow that integrated code, graphics, and informative text was made possible by this configuration, which enabled interactive data exploration, analysis, and documentation.
    
    \item \textbf{Python Libraries}:
    \begin{itemize}
        \item \textbf{XGBoost}: For training and assessing gradient-boosted models tailored for unbalanced datasets, this package was used. XGBoost was crucial for fraud detection tasks because of its effectiveness with high-dimensional data and its capacity to handle class imbalance.
        
        \item \textbf{Imbalanced-learn (imblearn)}: Sampling techniques such as SMOTE and random under-sampling were applied to manage the significant class imbalance in the data set. Imbalanced-learn provided robust tools for these sampling techniques, which were crucial for improving the model’s sensitivity to minority classes (fraudulent transactions).
        
        \item \textbf{CTGAN}: CTGAN was utilized to create synthetic data for the minority class in order to further solve the data imbalance. The variety of the dataset was improved by CTGAN's capacity to generate realistic synthetic samples, which enabled more precise model training.
        
        \item \textbf{Scikit-learn}:A full range of tools for evaluating models, such as precision, recall, F1-score, and other metrics and StandardScaler for preprocessing data, were provided by Scikit-learn. These measures were essential for evaluating the performance of the model and scaling the data, especially when it came to an unbalanced classification setting like fraud detection.
    \end{itemize}
\end{itemize}

\section{Dataset}

The dataset used in this study is the publicly available credit card fraud detection dataset from Kaggle \cite{dalpozzolo2016credit}. This dataset contains a total of 284,807 credit card transactions made by European cardholders over a two-day period. Among these transactions, only 0.172\% are labeled as fraudulent, creating a highly imbalanced dataset with a significant class disparity between fraudulent and legitimate transactions.

Each transaction in the dataset is represented by 30 features, including 28 anonymized variables denoted as \texttt{V1} to \texttt{V28}, and two additional attributes: \texttt{Time} and \texttt{Amount}. The anonymized features were generated through a principal component analysis (PCA) transformation to protect user confidentiality, and their exact meanings remain undisclosed. The \texttt{Time} feature captures the seconds elapsed between each transaction and the first transaction in the dataset, while \texttt{Amount} represents the monetary value of the transaction.

The target variable, \texttt{Class}, indicates the transaction type, where a value of 0 corresponds to a legitimate transaction and 1 represents a fraudulent transaction. Due to the severe imbalance in the dataset, with fraudulent transactions being exceedingly rare, the performance of the machine learning model must be evaluated with special attention to metrics that emphasize its ability to correctly identify fraud cases.

The training data was subjected to a number of resampling procedures in order to rectify the imbalance, guaranteeing that the model maintains its emphasis on the minority class without sacrificing generality.

\section{Dataset Analysis and Visualization}

To better understand the distribution of classes in the dataset, we analyzed the number of legitimate and fraudulent transactions.
\subsection{Class Imbalance}
\begin{figure}[H]
    \centering
    \includegraphics[width=0.7\textwidth]{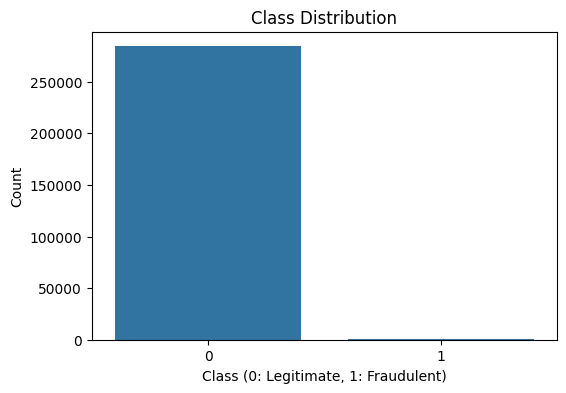}
    \caption{Class distribution of the credit card transactions dataset, showing a stark imbalance between legitimate (Class 0) and fraudulent (Class 1) transactions.}
    \label{fig:class_imbalance}
\end{figure}

As shown in Figure~\ref{fig:class_imbalance}, the dataset is highly imbalanced, with only a small fraction of transactions labeled as fraudulent.

This class imbalance presents a challenge for traditional machine learning algorithms, as the model may become biased towards predicting the majority class (legitimate transactions). Therefore, specialized techniques are applied to improve the model’s sensitivity to fraudulent transactions, ensuring more accurate detection.

\subsection{Transaction Amount by Class}

\begin{figure}[H]
    \centering
    \includegraphics[width=0.7\textwidth]{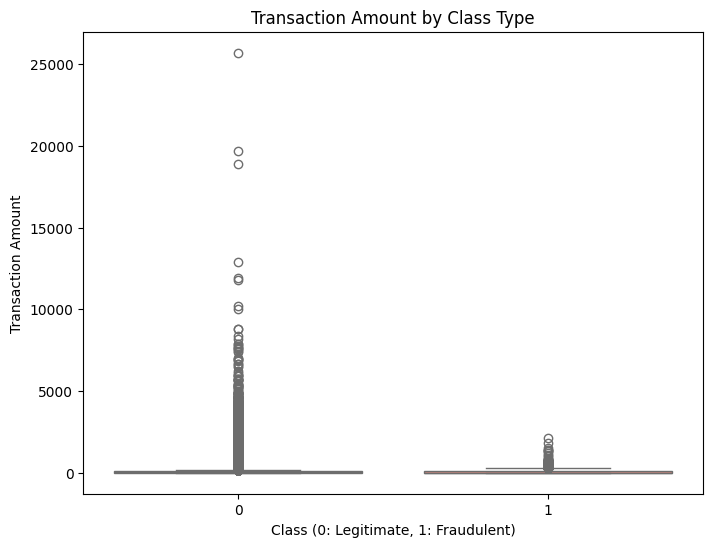}
    \caption{Boxplot of transaction amounts by class type, showing the distribution and range of amounts for legitimate (Class 0) and fraudulent (Class 1) transactions. Legitimate transactions exhibit a wider range, including high-value outliers, whereas fraudulent transactions tend to be lower in value.}
    \label{fig:amount_boxplot}
\end{figure}

The distribution of transaction amounts for legitimate and fraudulent transactions is illustrated in Figure~\ref{fig:amount_boxplot}. The boxplot shows that most transactions, both legitimate (Class 0) and fraudulent (Class 1), cluster around lower amounts. However, legitimate transactions have a wider range, with several high-value outliers extending beyond $25,000$, while fraudulent transactions tend to be concentrated at lower values with very few high-value outliers.

This distribution suggests that fraudsters may deliberately avoid high transaction amounts, possibly to evade detection mechanisms that are more likely to scrutinize large transactions. By keeping transaction values low, fraudsters might aim to blend in with typical transaction patterns and reduce the risk of triggering alerts. This insight could be valuable when combining transaction amount with other features to build a more robust detection model.

\subsection{Feature Correlations}

\begin{figure}[H]
    \centering
    \includegraphics[width=0.7\textwidth]{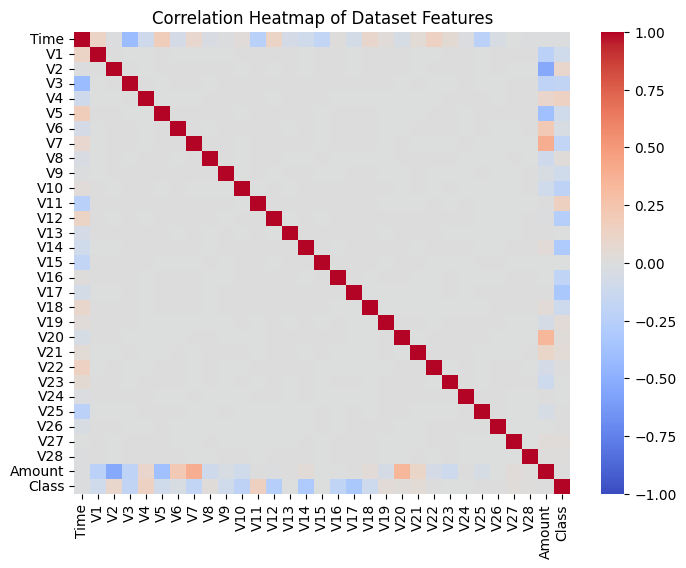}
    \caption{Correlation heatmap of features in the dataset, showing relationships between variables.}
    \label{fig:correlation_heatmap}
\end{figure}

A correlation heatmap (Figure~\ref{fig:correlation_heatmap}) was created to understand relationships among the anonymized features. This analysis can reveal potential interactions between features, aiding in feature selection and model interpretation.

\subsection{Distribution of Features in Data Set}

The distribution of features in the dataset is crucial for detecting patterns and differences between legitimate and fraudulent transactions. By analyzing the distribution of each feature, we can identify characteristics that might distinguish fraudulent transactions from legitimate ones, which can be beneficial for building effective detection models.

\begin{figure}[H]
    \centering
    \includegraphics[width=0.85\textwidth]{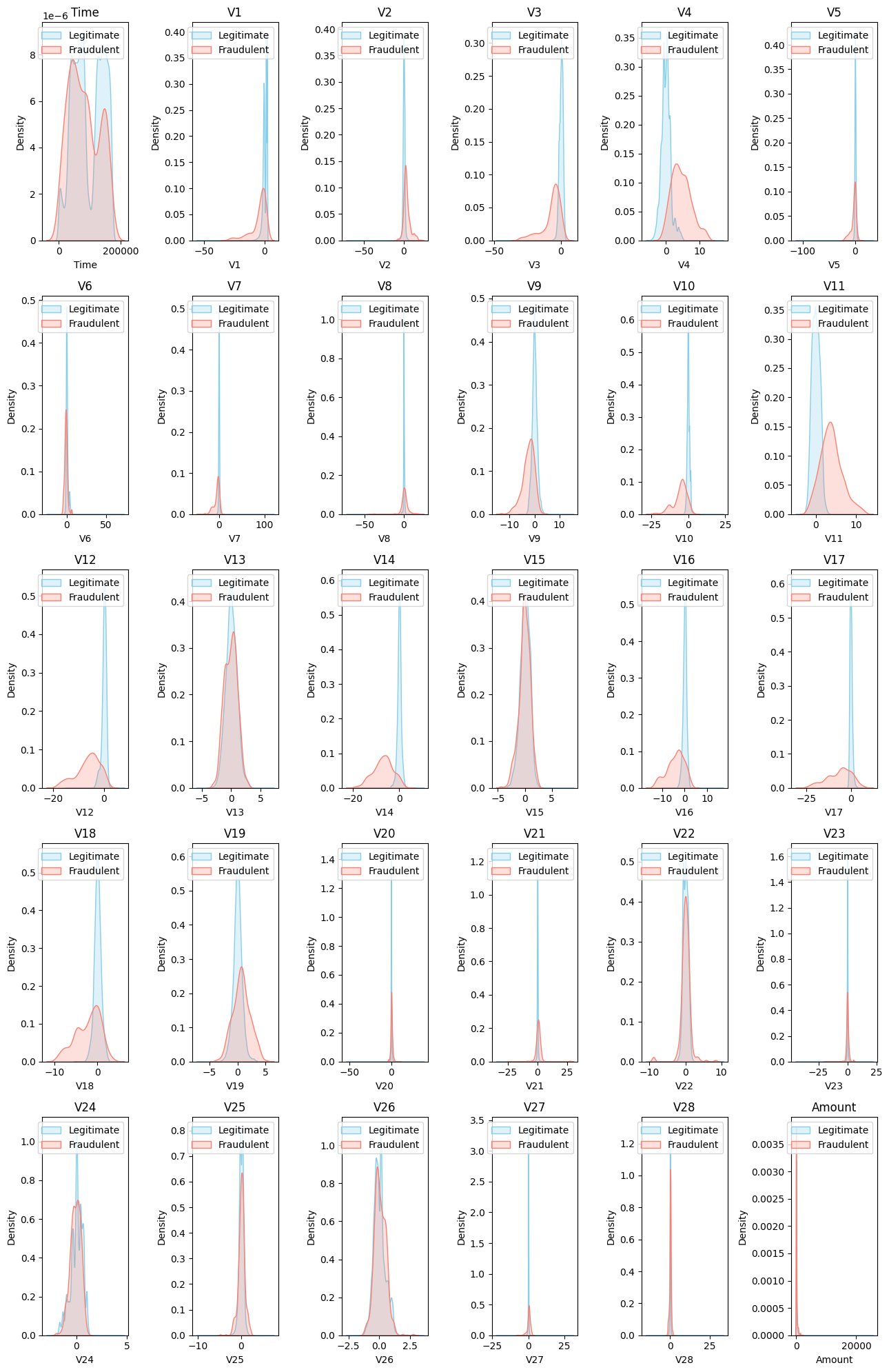}
    \caption{Distribution of selected features in the dataset, comparing legitimate and fraudulent transactions. Features like \texttt{V1}, \texttt{V2}, and \texttt{V3} show distinct patterns between classes, which could aid in fraud detection.}
    \label{fig:all_features_distribution}
\end{figure}

Figures~\ref{fig:all_features_distribution} shows the distributions of all features, for both legitimate and fraudulent transactions. Each plot provides a side-by-side comparison, enabling a clearer view of how these features vary across transaction types. Some features exhibit distinct patterns between classes, which may be valuable for the model to identify fraudulent behavior.

\section{Evaluation Metrics}
The algorithms' ability to identify fraudulent transactions was evaluated using a number of assessment indicators. The dataset's extreme class imbalance means that typical measurements might not give a complete picture of the model's effectiveness. To guarantee a thorough assessment of the models' performance, we used a variety of measures, such as Recall, Precision, F1-score, Matthews Correlation Coefficient (MCC), Accuracy (ACC), and Area Under the Receiver Operating Characteristic Curve (AUC).

\subsection{Recall}
Recall, measures the proportion of actual fraudulent transactions correctly identified by the model. This metric is particularly important in fraud detection, where it is crucial to minimize the number of undetected fraud cases. Recall is defined as:
\begin{equation}
    \text{Recall} = \frac{\text{True Positives}}{\text{True Positives} + \text{False Negatives}}
\end{equation}

\subsection{Precision}
Precision quantifies the proportion of transactions predicted as fraudulent that are indeed fraudulent. High precision indicates that the model is effective at minimizing false positives, thus reducing the number of legitimate transactions incorrectly flagged as fraud. Precision is given by:
\begin{equation}
    \text{Precision} = \frac{\text{True Positives}}{\text{True Positives} + \text{False Positives}}
\end{equation}

\subsection{F1-score}
The F1-score is the harmonic mean of Precision and Recall, providing a single metric that balances these two measures. This score is particularly useful when there is a significant class imbalance, as it considers both false positives and false negatives. The F1-score is calculated as:
\begin{equation}
    \text{F1-score} = 2 \times \frac{\text{Precision} \times \text{Recall}}{\text{Precision} + \text{Recall}}
\end{equation}

\subsection{Matthews Correlation Coefficient (MCC)}
A balanced statistic that accounts for both true and false positives as well as negatives, the Matthews Correlation Coefficient (MCC) offers a thorough assessment of model performance even when there are imbalanced classes present. Perfect predictions are represented by an MCC of +1, no better performance than chance is represented by an MCC of 0, and total misclassification is represented by an MCC of -1. MCC is determined as:
\begin{equation}
    \text{MCC} = \frac{\text{(TP} \times \text{TN)} - (\text{FP} \times \text{FN})}{\sqrt{(\text{TP} + \text{FP})(\text{TP} + \text{FN})(\text{TN} + \text{FP})(\text{TN} + \text{FN})}}
\end{equation}

\subsection{Accuracy (ACC)}
Accuracy measures the proportion of correctly classified transactions among all transactions. While useful, accuracy may not be fully indicative of performance in imbalanced datasets, as it could be dominated by the majority class. Accuracy is defined as:
\begin{equation}
    \text{Accuracy} = \frac{\text{True Positives} + \text{True Negatives}}{\text{Total Samples}}
\end{equation}

\subsection{Area Under the Curve (AUC)}
The Area Under the Receiver Operating Characteristic Curve (AUC) provides a measure of the model’s ability to discriminate between classes across different threshold levels. A higher AUC value indicates a better-performing model, with an AUC of 1 representing perfect classification and 0.5 representing random guessing.

These metrics provide a comprehensive evaluation of the models' performance, particularly in the context of detecting fraudulent transactions within an imbalanced dataset. Each metric captures a unique aspect of the model's effectiveness, aiding in a thorough comparison of different sampling techniques and model configurations.

\section{Results and Discussion}

In this section, we examine the performance of the XGBoost model in handling class imbalance, comparing results across different data imbalance handling techniques. The focus is on evaluating the model's effectiveness when sampling is applied before the train-test split and then comparing it to other models.

\subsection{Sampling Before Train-Test Split}

In this subsection, we analyze the results of applying sampling techniques before the train-test split. While this approach can improve model performance metrics, it risks data leakage, which may lead to artificially inflated evaluation scores. Table \ref{tab:xgboost_results} presents the performance metrics of the XGBoost model under this condition.

\begin{table}[h!]
\centering
\caption{XGBoost Performance Metrics with SMOTE Sampling Before Train-Test Split}
\label{tab:xgboost_results}
\begin{tabular}{|c|c|c|c|c|c|}
\hline
\textbf{Metric} & \textbf{Accuracy (\%)} & \textbf{Precision (\%)} & \textbf{Recall (\%)} & \textbf{F1-score (\%)} & \textbf{AUC (\%)} \\ \hline
\textbf{XGBoost} & 99.969 & 99.938 & 100 & 99.969 & 99.969 \\ \hline
\end{tabular}
\end{table}

As shown in Table \ref{tab:xgboost_results}\cite{9719580}, the XGBoost model achieves near-perfect scores in all metrics, with an accuracy of 99.969\%, precision of 99.938\%, recall of 100\%, F1-score of 99.969\%, and AUC of 99.969\%. However, the near-perfect performance is likely influenced by data leakage due to applying sampling before the train-test split. This issue highlights the importance of proper sampling technique implementation to ensure reliable model evaluation, as pre-split sampling can lead to overly optimistic results that may not generalize well to new data.

We shall now also look at the performance of an Autoencoder-enhanced XGBoost model with SMOTE and CGAN synthetic data generation (AE-XGB-SMOTE-CGAN). Table \ref{tab:ae_xgb_smote_cgan_results} provides the performance metrics for AE-XGB-SMOTE-CGAN with a decision threshold of 0.05, capturing accuracy, true positive rate (TPR), true negative rate (TNR), and Matthews Correlation Coefficient (MCC).

\begin{table}[h!]
\centering
\caption{Performance of AE-XGB-SMOTE-CGAN Model at Threshold = 0.05}
\label{tab:ae_xgb_smote_cgan_results}
\begin{tabular}{|c|c|c|c|c|}
\hline
\textbf{Model} & \textbf{ACC (\%)} & \textbf{TPR (\%)} & \textbf{TNR (\%)} & \textbf{MCC} \\ \hline
AE-XGB-SMOTE-CGAN (threshold = 0.05) & 99.87 & 89.29 & 99.32 & 0.773 \\ \hline
\end{tabular}
\end{table}

While this is indeed a novel approach as the authors claim \cite{Du2024} and shows great prediction capabilities, the timing of the sampling does rise concerns regards to data leakage.   

These results emphasize the need for careful application of sampling methods, particularly in imbalanced datasets, where the risk of data leakage can significantly distort model evaluation. The following sections will compare these findings with results obtained using post-split sampling and other classifiers.

\subsection{Sampling After Train-Test Split}

For this section we will first look into results from our own experiments:

\begin{table}[h!]
\centering
\resizebox{\textwidth}{!}{%
\begin{tabular}{|l|c|c|c|c|}
\hline
\textbf{Model} & \textbf{Accuracy} & \textbf{Precision} & \textbf{Recall} & \textbf{F1 Score} \\ \hline
Baseline model (No imbalance handling) & 99.96\% & 96.50\% & 91.50\% & 94.00\% \\ \hline
Baseline model with Random Over-Sampling & 99.95\% & 92.50\% & 92.00\% & 92.50\% \\ \hline
Baseline model with ROS and cost-sensitive learning & 99.95\% & 93.50\% & 93.00\% & 93.10\% \\ \hline
Baseline with SMOTE & 99.95\% & 93.00\% & 92.50\% & 92.56\% \\ \hline
Baseline with SMOTE and Random Under Sampling pipeline & 99.95\% & 93.75\% & 93.00\% & 93.50\% \\ \hline
Baseline with regularization(reg\_alpha=0.6, reg\_lambda=0.2) & 99.96\% & 97.50\% & 91.50\% & 94.30\% \\ \hline
Baseline with CGAN and Cost Sensitive Learining(scaleposweight=577.27/10)
 & 99.97\%& 97.00\% & 93.00\% & 95.00\% \\ \hline
\end{tabular}%
}
\caption{Performance metrics for various sampling techniques}
\label{tab:model_performance}
\end{table}

As we will see these results are quite descent results compared to Random Forest and aren't really that far either from the results of the previously shown models which did sampling pre the train-test split.

The parameters of the above baseline model are as follows:

\begin{table}[h!]
    \centering
     \begin{tabular}{|>{\raggedright\arraybackslash}p{4cm}|>{\raggedright\arraybackslash}p{5cm}|}
        \hline
        \textbf{Parameter} & \textbf{Value} \\
        \hline
        use\_label\_encoder & False \\
        \hline
        eval\_metric & logloss \\
        \hline
        random\_state & 42 \\
        \hline
        learning\_rate & 0.4 \\
        \hline
        n\_estimators & 1000 \\
        \hline
        tree\_method & hist \\
        \hline
        n\_jobs & -1 \\
        \hline
        device & gpu \\
        \hline
        objective & binary:logistic \\
        \hline
    \end{tabular}
    \caption{Baseline XGBoost Model Parameters}
\end{table}

Moving on to results by other authors, we can also see some impressive scores with no imbalance handling from the paper by K.K. Mohbey\cite{Mohbey2022}:
\begin{table}[H]
\centering
\begin{tabular}{|l|c|c|c|c|}
\hline
\textbf{Model} & \textbf{Accuracy} & \textbf{Precision} & \textbf{Recall} & \textbf{F1 Score} \\ \hline
XGBoost & 96.44\% & 96.00\% & 97.00\% & 96.00\% \\ \hline
\end{tabular}
\caption{Performance metrics for the XGBoost model}
\label{tab:xgboost_performance}
\end{table}

For a comparison with Random Forest we can look at Chougugudza's work\cite{Chogugudza2022}, where he applied SMOTE pre train-test split to improve the classifier:

\begin{table}[h!]
\centering
\begin{tabular}{|l|c|c|c|c|}
\hline
\textbf{Model} & \textbf{Accuracy} & \textbf{Precision} & \textbf{Recall} & \textbf{F1 Score} \\ \hline
Random Forest & 99.95\% & 95.00\% & 91.00\% & 92.50\% \\ \hline
\end{tabular}
\caption{Performance metrics for the Random Forest model}
\label{tab:randomforest_performance}
\end{table}

From these results it is now evident that the xgboost model is still superior even without the risk of data leakage which leads to inflated metric scores. 

\section*{Conclusion}
This study showed that the XGBoost model can detect credit card fraud with high performance without using pre-split sampling, which poses a serious danger of data leaking. We avoided falsely inflated metrics, which are frequently the result of inappropriate sampling processes, and maintained the integrity of model assessment by concentrating on post-split sampling techniques.

Our results confirm that XGBoost retains both precision and resilience in detecting fraudulent transactions in extremely unbalanced datasets when used in conjunction with well-executed post-split sampling. These findings support the usefulness of XGBoost as an effective fraud detection technique that produces dependable results without sacrificing authenticity. Alternative sampling strategies that improve model sensitivity to minority classes while maintaining strict data leakage protections should be further explored in future research.

\printbibliography

\newpage
\appendix
\section*{Appendix A: SMOTE and Random Under-Sampling Pipeline}

\lstset{
    language=Python,
    basicstyle=\ttfamily\small,
    keywordstyle=\color{blue},
    commentstyle=\color{gray},
    stringstyle=\color{orange},
    backgroundcolor=\color{lightgray!20},
    frame=single,
    breaklines=true,
    showstringspaces=false,
    tabsize=4,
    captionpos=b,
    numbers=left,
    numberstyle=\tiny\color{gray}
}

\begin{lstlisting}[caption={Python code for SMOTE and Random Under-Sampling Pipeline}]
from imblearn.under_sampling import RandomUnderSampler
from imblearn.pipeline import Pipeline as ImbPipeline

# Defining over-sampling and under-sampling strategies
over = SMOTE(sampling_strategy=0.8, random_state=RANDOM_STATE)
under = RandomUnderSampler(sampling_strategy=0.9, random_state=RANDOM_STATE)

# Create a pipeline, also consider u -> o
steps = [('o', over), ('u', under)]
pipeline = ImbPipeline(steps=steps)

X_train_resampled, y_train_resampled = pipeline.fit_resample(X_train, y_train)
\end{lstlisting}

\newpage

\section*{Appendix B: Generating Synthetic Data Using CGAN}

\lstset{
    language=Python,
    basicstyle=\ttfamily\small,
    keywordstyle=\color{blue},
    commentstyle=\color{gray},
    stringstyle=\color{orange},
    backgroundcolor=\color{lightgray!20},
    frame=single,
    breaklines=true,
    showstringspaces=false,
    tabsize=4,
    captionpos=b,
    numbers=left,
    numberstyle=\tiny\color{gray}
}

\begin{lstlisting}[caption={Python code for generating synthetic data using CGAN}]
# Combine X_train and y_train for CTGAN model training
train_data1 = pd.concat([X_train, y_train], axis=1)

# Separate minority class data from training set
minority_class_data1 = train_data1[train_data1['Class'] == 1]

ctgan1 = CTGAN(epochs=300)
ctgan1.fit(minority_class_data1)

# Generate synthetic samples for the minority class
synthetic_data1 = ctgan1.sample(227057)
synthetic_data1['Class'] = 1

# Combine synthetic data with the original training data
train_data_balanced1 = pd.concat([train_data1, synthetic_data1], ignore_index=True)

# Separate features and labels for the balanced training data
X_train_balanced1 = train_data_balanced1.drop('Class', axis=1)
y_train_balanced1 = train_data_balanced1['Class']
\end{lstlisting}

\newpage 

\section*{Appendix C: Data Preprocessing for Experiments}

\lstset{
    language=Python,
    basicstyle=\ttfamily\small,
    keywordstyle=\color{blue},
    commentstyle=\color{gray},
    stringstyle=\color{orange},
    backgroundcolor=\color{lightgray!20},
    frame=single,
    breaklines=true,
    showstringspaces=false,
    tabsize=4,
    captionpos=b,
    numbers=left,
    numberstyle=\tiny\color{gray}
}

\begin{lstlisting}[caption={Python code for data preprocessing on our experiments}]
scaler_amount = StandardScaler()
data['Amount_Scaled'] = scaler_amount.fit_transform(data['Amount'].values.reshape(-1, 1))

# Include the 'Time' feature and scale it
scaler_time = StandardScaler()
data['Time_Scaled'] = scaler_time.fit_transform(data['Time'].values.reshape(-1, 1))

# Feature Engineering: Create 'Hour' feature from 'Time'
data['Hour'] = (data['Time'] // 3600) % 24

# Creating a 'Day_Segment' feature from the 'Time' column...
def assign_day_segment(hour):
    if 6 <= hour < 12:
        return 'Morning'
    elif 12 <= hour < 18:
        return 'Afternoon'
    elif 18 <= hour < 24:
        return 'Evening'
    else:
        return 'Night'

data['Day_Segment'] = data['Hour'].apply(assign_day_segment)

# One-hot encode 'Day_Segment'
data = pd.get_dummies(data, columns=['Day_Segment'], drop_first=True)

# Drop original Amount and Time cols
data = data.drop(['Amount','Time'], axis=1)

# Rearranging columns to place new features at the end
cols = data.columns.tolist()
cols = [col for col in cols if col != 'Class'] + ['Class']
data = data[cols]

# Separate features and target variable
X = data.drop('Class', axis=1)
y = data['Class']
\end{lstlisting}

\end{document}